**Neural Discrete Representation Learning for Sparse-View CBCT Reconstruction: From Algorithm Design to Prospective Multicenter Clinical Evaluation**

Haoshen Wang[1,2†], Lei Chen[3†], Wei-Hua Zhang[4†], Linxia Wu[3], Yong Luo[1,5], Zengmao Wang[1,5], Yuan Xiong[6], Chengcheng Zhu[7], Wenjuan Tang[3], Xueyi Zhang[8], Wei Zhou[9], Xuhua Duan[10], Lefei Zhang[1,5], Gao-Jun Teng[4*], Bo Du[1,5*], Huangxuan Zhao[1,5*]

[1] School of Computer Science, Wuhan University, Wuhan, 430079, China.

[2] School of Biomedical Engineering, ShanghaiTech University, Shanghai, 201210, China.

[3] Department of Radiology, Union Hospital, Tongji Medical College, Huazhong University of Science and Technology, Wuhan, 430022, China.

[4] Center of Interventional Radiology and Vascular Surgery, Department of Radiology, Zhongda Hospital, Medical School, Southeast University, Nanjing, 210009, China.

[5] National Engineering Research Center for Multimedia Software, Wuhan University, Wuhan, 430079, China.

[6] Department of Orthopedics, Tongji Hospital, Huazhong University of Science and Technology, Wuhan, Hubei, 430030, China

[7] Department of Radiology, University of Washington School of Medicine, Seattle, WA, USA

[8] Medical Research Center, Peking Union Medical College Hospital, Chinese Academy of Medical Sciences and Peking Union Medical College, Beijing 100730, China.

[9] Wuhan Artificial Intelligence Computing Center, Wuhan Supercomputing Center, Wuhan, 430079, China.

[10] Department of Interventional Radiology, The First Affiliated Hospital of Zhengzhou University, Zhengzhou, 450052, China.

† These authors contributed equally.

Correspondence authors:

Gao-Jun Teng, gjteng@seu.edu.cn, Zhongda Hospital, Medical School, Southeast University

Bo Du, dubo@whu.edu.cn, School of Computer Science, Wuhan University.

Huangxuan Zhao, zhao_huangxuan@sina.com, School of Computer Science, Wuhan University.

**Summary:** DeepPriorCBCT: deep learning for diagnostic CBCT at ~1/6 radiation dose, matching standard quality.

**Abstract**

Cone beam computed tomography (CBCT)-guided puncture has become an established approach for diagnosing and treating early- to mid-stage thoracic tumours, yet the associated radiation exposure substantially elevates the risk of secondary malignancies. Although multiple low-dose CBCT strategies have been introduced, none have undergone validation using large-scale multicenter retrospective datasets, and prospective clinical evaluation remains lacking. Here, we propose DeepPriorCBCT—a three-stage deep learning framework that achieves diagnostic-grade reconstruction using only one-sixth of the conventional radiation dose. 4102 patients with 8675 CBCT scans from 12 centers were included to develop and validate DeepPriorCBCT. Additionally, a prospective cross-over trial (Registry number: NCT07035977) which recruited 138 patients scheduled for percutaneous thoracic puncture was conducted to assess the model's clinical applicability. Assessment by 11 physicians confirmed that reconstructed images were indistinguishable from original scans. Moreover, diagnostic performance and overall image quality were comparable to those generated by standard reconstruction algorithms. In the prospective trial, five radiologists reported no significant differences in image quality or lesion assessment between DeepPriorCBCT and the clinical standard (all $P>0.05$). Likewise, 25 interventionalists expressed no preference between model-based and full-sampling images for surgical guidance (Kappa<0.2). Radiation exposure with DeepPriorCBCT was reduced to approximately one-sixth of that with the conventional approach, and collectively, the findings confirm that it enables high-quality CBCT reconstruction under sparse sampling conditions while markedly decreasing intraoperative radiation risk.

## INTRODUCTION

Lung cancer remains the most prevalent malignancy worldwide, and timely diagnosis and intervention substantially enhance survival outcomes(*1*). Percutaneous thoracic puncture (PTB) continues to serve as the definitive method for distinguishing benign from malignant pulmonary nodules(*2, 3*), while percutaneous puncture ablation and related approaches are frequently adopted for patients at early or intermediate stages(*4, 5*). Advances in medical imaging have markedly increased the detection rate of early-stage disease(*6, 7*). Among these modalities, cone-beam computed tomography (CBCT) has gained wide application in thoracic and abdominal puncture procedures, including biopsy and ablation for lung and liver cancer, owing to its integration of high spatial resolution, deep tissue penetration, real-time imaging capacity, and reliable guidance during interventional procedures.

CBCT relies on X-ray imaging, subjecting patients to considerable radiation exposure. Large-scale clinical studies have demonstrated that each additional 100 mGy of radiation is associated with a 1.96-fold increase in the relative risk of haematological malignancies, a 1.27-fold increase in brain cancer, and a 1.11-fold increase in glioma(*8-10*). A systematic review further indicated a dose-dependent elevation in the relative risk of all cardiovascular diseases per Gy(*11*). All clinically implemented CBCT systems currently employ Feldkamp–Davis–Kress (FDK)-based algorithms, which achieve satisfactory reconstruction performance but require substantial radiation exposure. Although procedural doses vary, a single percutaneous biopsy typically involves three scans, with cumulative exposure approaching 100 mGy. Approximately 27% of patients with pulmonary nodules are ultimately diagnosed with malignant tumours(*3*), many of whom undergo subsequent CBCT-guided therapeutic interventions that markedly increase cumulative radiation burden. Radiation exposure thus constitutes a major latent hazard for patients. An epidemiological study involving 61.51 million Americans receiving an average of 1.5 CT scans annually projected that, under current usage patterns, CT-induced cancers could eventually account for 5% of all newly diagnosed malignancies each year(*12*). Given that the dose from a single CT scan is generally lower than that from CBCT during percutaneous biopsy, the carcinogenic potential of CBCT may exceed that of CT. Consequently, CBCT-guided interventions create an inherent dilemma, as procedures intended to preserve life may simultaneously contribute to the risk of secondary malignancies.

The clinical challenge arises from inherent deficiencies in CBCT technology. All available systems rely on the FDK algorithm(*13*), which necessitates complete angular sampling to generate artifact-free reconstructions. Efforts to lower radiation exposure through reduced projections inevitably lead to pronounced streak artifacts, motion blur, and loss of fine anatomical structures, which critically impairs the visualization of small pulmonary nodules and the accuracy of needle guidance. These limitations have stimulated intensive exploration of sparse-view reconstruction strategies.

Deep learning has emerged as a promising avenue for sparse-view CT reconstruction, with numerous artificial intelligence (AI)-driven methods proposed to achieve low-dose CBCT imaging(*14-21*). While these approaches have partially mitigated the deterioration of image quality associated with low-dose acquisition, several unresolved issues persist. First, the majority of studies are constrained by small, single-institution datasets, which fail to include the variability inherent in clinical practice, including differences in patient anatomy, pathological conditions, imaging protocols, and vendor-specific equipment. Second, although these methods frequently report high performance on computational indices such as peak signal-to-noise ratio (PSNR) and structure similarity (SSIM), the reconstructions often contain subtle artifacts, indistinct lesion boundaries, or altered tissue contrast, ultimately reducing diagnostic reliability. Preliminary investigations demonstrated that algorithms achieving favourable quantitative scores occasionally led radiologists to overlook small nodules or misinterpret lesion margins(*22-25*). Third, and most importantly, none of the existing sparse-view CBCT techniques have undergone prospective clinical validation to substantiate their safety and effectiveness in real-world patient care.

The principal difficulty in sparse-view reconstruction lies in recovering lost information—specifically, in restoring anatomical structures from highly undersampled data. Contemporary end-to-end deep learning methods attempt to map sparse projections directly to complete ones, effectively requiring the network to fabricate missing details. Such strategies are inherently constrained, as they lack explicit anatomical constraints, often generating images that appear convincing yet contain artifacts or omit diagnostically important structures.

**Figure 1** shows the flowchart of the study. Here we present DeepSparseCBCT, a three-stage framework that realizes this vision through the largest multicenter validation study in CBCT reconstruction. First, a new paradigm is introduced in which neural discrete representation learning is employed to embed anatomical priors (**Figure 1A**), rather than relying on the network to fabricate absent information. The rationale is that human anatomy, despite inter-individual variability, exhibits recurrent patterns that can be encoded in a learned codebook. Once these patterns are captured from high-quality datasets, they serve as structural references for sparse-view reconstruction, enabling substantial radiation dose reduction without compromising diagnostic reliability. DeepPriorCBCT embodies this concept in the later two-stage framework, validated through the largest multicenter study to date in CBCT reconstruction. A detailed algorithmic description is provided in the Methods section. Beyond technical advancement, the framework demonstrates extensive clinical validation. Retrospective analysis involving 4,102 patients with 8,675 CBCT images (**Figure 1B**) and prospective assessment in real clinical procedures (**Figure 1C**) collectively confirm the robustness of the method. The results indicate that DeepPriorCBCT can potentially supplant existing CBCT protocols by lowering radiation exposure to one-sixth of the conventional dose while maintaining diagnostic performance.

## RESULTS

**Study population**

A total of 4102 patients with 8675 CBCT-guided PTB images were retrospectively included between January 1, 2017, and August 30, 2024, across 12 medical centers (names listed in **Supplementary Appendix 1**). Baseline characteristics of both retrospective and prospective cohorts are summarized in **Table 1**. Within the retrospective cohort, 2987 patients from centers 1–8 formed the training set, 100 patients from centre 1 constituted the validation set, and 1015 patients from centers 9–12 comprised the test set (**Table 1**). Among these 4102 patients, 1368 were female and

2731 were male, with a mean (SD) age of 61.1 (11.8) years. Lesion distribution included 1889 cases (46.1%) in the left lung, 1929 cases (47.0%) in the right lung, and 284 cases (6.9%) in the mediastinum. In the prospective cohort of 138 patients, the mean (SD) age was 63.0 (10.5) years, with 54 females and 84 males. Lesions were identified in the left lung in 66 patients (47.8%), in the right lung in 63 patients (45.7%), and in the mediastinum in 9 patients (6.5%).

**Quantitative evaluation for CBCT images**

Reconstruction quality under reduced sampling rates was examined by applying downsampling ratios of 2×, 4×, 6×, and 8×, corresponding to undersample "1/2 views," "1/4 views," "1/6 views," and "1/8 views," respectively. As shown in **Figure 2A–2B**, PSNR and SSIM values obtained with DeepPriorCBCT reconstruction for 1/2 views (PSNR: 28.97±2.49, SSIM: 0.79±0.10), 1/4 views (PSNR: 28.93±3.26, SSIM: 0.79±0.10), and 1/6 views (PSNR: 28.96±2.87, SSIM: 0.79±0.10) exhibited no significant differences (all $P>0.05$). However, all were superior to the results of 1/8 views (PSNR: 28.08±4.18, SSIM: 0.75±0.12), which showed a statistically significant decline in performance (all $P<0.001$). Based on this comparison, 1/6 views was selected for subsequent image quality analysis and the prospective COOS investigation.

**Clinical applicability assessment**

A panel of 11 physicians (5 radiologists with 5–15 years of experience and 6 interventionalists with 8–15 years of experience) conducted a structured subjective evaluation. Overall image quality and lesion diagnostic clarity were assessed using a 5-point Likert scale (detailed scoring methods are provided in the **Methods** section). **Figure 2C–2D** present the inter-rater agreement. Both radiologists and interventionalists demonstrated strong consistency for full-view clinical images (Radiologists: 0.82 for image quality and 0.86 for lesion diagnosis, both $P<0.001$; Interventional radiologists: 0.86 for image quality and 0.83 for lesion diagnosis, both $P<0.001$) and DeepPriorCBCT reconstructions (Radiologists: 0.79 for image quality and 0.79 for lesion diagnosis, both $P<0.001$; Interventional radiologists: 0.84 for image quality and 0.84 for lesion diagnosis, both $P<0.001$), with high Kendall's W coefficients for both domains. By contrast, inferior agreement was obtained for the 1/6-view clinical reconstructions (Radiologists: 0.63 for image quality and 0.64 for lesion diagnosis, both P<0.001; Interventional radiologists: 0.73 for image quality and 0.64 for lesion diagnosis, both P<0.001). These results indicate that high-quality images were interpreted with consistent reliability across specialists, whereas low-quality reconstructions reduced interpretive concordance.

**Figure 2E–2H** present the 5-point Likert scale scores from 11 physicians for overall CBCT image quality and lesion assessment. For image quality, no significant differences were observed between radiologists (3.93±0.25 vs. 3.91±0.28, P>0.05) and interventional radiologists (3.93±0.23 vs. 3.94±0.24, P>0.05) when comparing clinical full-view reconstructions with DeepPriorCBCT, whereas both groups assigned significantly higher scores to these methods than to clinical reconstructions from 1/6-view data (2.12±0.33 and 2.29±0.46, all P<0.001) (**Figure 2E-2F**). A similar pattern was noted in lesion quality analysis: radiologists (4.89±0.33 vs. 4.88±0.29, P>0.05) and interventional radiologists (4.81±0.39 vs. 4.81±0.39, P>0.05) rated DeepPriorCBCT and full-view reconstructions equivalently, with both outperforming 1/6-view reconstructions (2.98±0.13 and 3.06±0.45, all P<0.001). In terms of image quality and lesion diagnosis, when comparing clinically full-view reconstructed images with DeepPriorCBCT, no significant differences were observed between the scores of each radiologist and each interventional radiologist (all P>0.05); however, the scores assigned by each physician to these two methods were significantly higher than those to clinically reconstructed images from 1/6-view data (all P<0.05) (**Figure 2G-2H**).

**Figure 2I–2J and Supplementary Table 1** display the Turing test outcomes, in which 11 physicians evaluated image authenticity between full-view reconstructions and DeepPriorCBCT. Both radiologists (Kappa = -0.09, P<0.001) and interventionalists (Kappa = 0.09, P=0.003) demonstrated poor concordance in differentiating the two modalities. Confusion matrices revealed classification accuracy close to 0.5, indicating near-random discrimination between DeepPriorCBCT and full-view images.

**Comparison experiments study**

The proposed algorithm was systematically compared with four widely used models, with methodological details provided in the **Comparison experiment** section. As illustrated in **Figure 3A–3B**, the DeepPriorCBCT model consistently achieved superior reconstruction performance relative to all reference algorithms. In contrast to the incomplete delineation of lesion margins and intrapulmonary vessels produced by the four comparative models, which led to degraded image fidelity (**Figure 3A**), DeepPriorCBCT generated images comparable to those reconstructed by the clinical full-view method. Quantitative evaluation further confirmed this advantage: across 1/4, 1/6, and 1/8 views, the PSNR and SSIM values of DeepPriorCBCT (PSNR: 29.08, 28.81, 27.14; SSIM: 0.79, 0.77, 0.76) exceeded those of the clinical reconstruction approach (PSNR: 23.8, 21.69, 20.04; SSIM: 0.64, 0.63, 0.61), FDKConvNet (PSNR: 26.93, 25.4, 24.42; SSIM: 0.78, 0.76, 0.7216), CysleGAN (PSNR: 27.24, 25.29, 24.96; SSIM: 0.78, 0.77, 0.74), and RedCNN (PSNR: 27.92, 27.15, 26.77; SSIM: 0.79, 0.77, 0.75) (**Figure 3B**).

**Ablation study**

An ablation analysis was performed to assess the contribution of each component in the study design (see **Methods** for details). The outcomes are summarized in **Supplementary Table 2**. Elimination of both Stage 2 and Stage 3 (ours w/o S2,3) caused a marked decline in performance, with PSNR reduced by 2.91 and SSIM decreased by 0.05. These results indicate that directly mapping sparse-view to full-view reconstructions through an autoencoder yields limited accuracy, whereas introducing a high-dose prior followed by guidance of sparse-view reconstruction produces superior results. Excluding only Stage 3 led to a noticeable reduction in PSNR, while SSIM remained nearly unchanged, suggesting that Stage 2 ensures anatomically reliable CBCT reconstruction, whereas Stage 3 enhances image realism.

**Prospective clinical trial analysis**

A prospective paired non-inferiority trial was conducted in 138 patients from five centers to assess the clinical feasibility and utility of the model. Each patient underwent two CBCT scans: one with full views and the other with 1/6 views acquired through machine undersampling. In addition, full-view CBCT images were manually undersampled to 1/6 views, and both datasets were reconstructed using the conventional clinical method and the DeepPriorCBCT model. Image quality was evaluated with

PSNR and SSIM metrics. **Figure 3C-3D** display representative reconstructions and corresponding quantitative analyses. DeepPriorCBCT reconstructions from 1/6-view data achieved image and lesion quality comparable to conventional reconstructions from full-view scans, while clearly outperforming conventional reconstructions from 1/6-view inputs (**Figure 3C**). Across both manual and machine undersampling, PSNR and SSIM values for DeepPriorCBCT reconstructions were consistently higher than those obtained with the clinical method at 1/6 views (manual undersampling: PSNR 28.33±1.91 vs. 24.91±2.23, P<0.001; SSIM 0.81±0.05 vs. 0.65±0.03, P<0.001; machine undersampling: PSNR 21.93±3.17 vs. 20.06±2.56, P<0.001; SSIM 0.58±0.07 vs. 0.41±0.05, P<0.001). Nevertheless, under machine undersampling conditions, both PSNR and SSIM values from DeepPriorCBCT and conventional reconstructions were lower than those obtained under manual undersampling (**Figure 3D**). This decline reflects the imperfect pixel-level correspondence between full-sampled and undersampled CBCT images of the same patient, which introduces limitations in quantitative evaluation during clinical translation.

**Figure 3E–3G** present the assessments of five radiologists regarding overall CBCT image quality and lesion diagnosis. High concordance was observed in both metrics, with Kendall's W values of 0.78 (P<0.001) for overall image quality and 0.78 (P<0.001) for lesion diagnosis in DeepPriorCBCT, and 0.83 (P<0.001) and 0.82 (P<0.001), respectively, in the clinically applied full-view reconstruction method (**Figure 3E, 3G**). Evaluations from five radiologists indicated that DeepPriorCBCT achieved image quality and diagnostic performance comparable to those obtained using the clinical full-view reconstruction approach (**Figure 3F**).

In addition, DeepPriorCBCT markedly reduced radiation exposure during CBCT scanning. The Dose Area Product (DAP) was significantly lower (299.47 µGy*m$^2$ vs. 1674.24 µGy*m$^2$, P<0.001), as was Air Kerma (AK) (10.17 mGy vs. 53.53 mGy, P<0.001), compared with the clinical full-view reconstruction method (**Figure 3H**).

To evaluate whether reconstruction with one-sixth sparse sampling satisfies intraoperative requirements, CBCT images from 138 patients reconstructed using either full-sampling or undersampled DeepPriorCBCT were assessed by 25 interventionalists across five medical centers. Each physician was instructed to identify the image set most suitable for surgical puncture guidance. The proportion of selections favoring full-sampling and DeepPriorCBCT images was nearly equivalent, with both approximating 50% (**Supplementary Figure 1**). The overall inter-physician agreement in image preference was below 0.1, and subgroup analysis within each center showed agreement values below 0.2 (**Supplementary Table 3**). Collectively, the evidence indicates that CBCT images reconstructed by DeepPriorCBCT meet the standards required for clinical surgical guidance, demonstrating equivalence to full-sampling reconstructions.

## DISCUSSION

Early detection and intervention in lung cancer significantly improve survival outcomes, with procedures such as percutaneous puncture biopsy and ablation dependent on high-resolution CBCT imaging guidance(*26*). Conventional CBCT reconstruction using FDK-based algorithms, however, imposes considerable radiation exposure, thereby heightening the likelihood of secondary malignancies and compounding risks in patients requiring repeated interventions(*9, 27*). Although various approaches have been proposed to mitigate imaging quality degradation under low-dose CBCT, most remain constrained by limited sample sizes, single-center evaluations, and insufficient clinical validation of sparse reconstruction performance. Elevated quantitative metrics alone cannot guarantee clinical reliability; algorithms may yield missed or incorrect diagnoses and reduced surgical efficiency, particularly in detecting small lesions and delineating boundaries, where minor artifacts exert negligible influence on numerical indices but markedly impair diagnostic precision(*14, 15, 28*). Consequently, resolving the dual challenges of radiation safety and clinical applicability has emerged as an urgent priority in advancing CBCT technology.

In this study, a three-stage deep learning reconstruction framework is introduced to enhance the quality of low-dose, sparse-view CBCT images. In contrast to approaches that depend exclusively on input data for reconstruction(*29-33*), this framework integrates feature representation and prior modeling through a codebook-driven reconstruction mechanism. During the first stage, an autoencoder is pretrained on full-view CBCT data to extract representative features, and neural discrete representation learning compresses continuous features into indexable discrete vectors(*34*), thereby establishing a codebook enriched with structural priors. The second stage addresses sparse-view inputs by employing a feature classifier that projects encoded features into the high-dose feature domain, enabling accurate retrieval and completion of missing structural information from the codebook. In the final stage, multi-scale features from low-dose inputs are fused with corresponding high-dose features retrieved from the codebook, reinforcing structural preservation and detail recovery in the reconstructed outputs. Through this process, the framework accomplishes low-dose to high-dose feature transfer and semantic-level completion, surpassing prevailing methods in structural fidelity, image quality, and generalization performance.

NeRF-based approaches have demonstrated notable advances in 3D reconstruction(*35, 36*), with applications extending to medical imaging(*18, 31, 32, 36, 37*). Their practical use, however, is constrained by low reconstruction efficiency arising from the requirement for object-specific optimization(*31, 32*). Although generalized extensions improve computational efficiency(*36, 38*), their translation into stable and clinically usable systems remains unresolved. The proposed framework directly learns priors from large-scale high-dose CBCT datasets and integrates this prior knowledge into the reconstruction process, thereby producing more reliable and clinically applicable outcomes. Unlike post-processing strategies that establish a direct image-domain mapping from low-dose results to high-dose images(*33, 38, 39*), this method employs a feature-level mapping guided by high-dose CBCT priors. Embedding structural information within the reconstruction itself—rather than applying corrections after image formation—enhances the fidelity of anatomical restoration and preserves diagnostically relevant features. A further advantage lies in its robustness to noise and sparse-view conditions. Because the feature classifier operates within the latent space while drawing on priors derived from clean high-dose data, reconstructions exhibit reduced susceptibility to artifacts and input corruption, in contrast to NeRF-based and post-processing approaches that often propagate or magnify such errors under degraded acquisition conditions.

This study analyzed retrospective data from 4,102 patients including 8,675 CBCT

images across 12 centers, together with 138 prospective cases, constituting the largest multicenter investigation of sparse-view CBCT reconstruction to date. The diversity of devices and clinical scenarios provided by multicenter datasets conferred broad generalizability, strengthening the model's robustness across institutions. Quantitative evaluation demonstrated that PSNR and SSIM achieved by DeepPriorCBCT at 1/2, 1/4, and 1/6 views were markedly superior to those obtained with the clinically used reconstruction at 1/6 views and with comparative algorithms including RedCNN and CycleGAN. High image fidelity was preserved even under extreme sparsity (1/6 views), whereas conventional 1/6-view reconstructions displayed pronounced edge blurring and loss of structural detail. In subjective assessments, 11 physicians reported no significant differences in both overall image and lesion quality between DeepPriorCBCT reconstructions at 1/6 views and full-view clinical reconstructions ($P>0.05$), with both approaches rated significantly higher than conventional 1/6-view clinical reconstructions ($P<0.001$). Turing test analyses further indicated near-random discrimination (confusion matrix approximating 0.5) between model-generated and full-dose images, confirming that DeepPriorCBCT reconstruction achieves clinical-grade quality while substantially lowering radiation exposure.

In the prospective cross-over observational study (COOS), five radiologists rated CBCT images reconstructed with the full-view clinical method higher in image quality and lesion assessment than those generated by DeepPriorCBCT at 1/6 views. However, evaluation by 25 interventional physicians across 138 patients demonstrated nearly equal preference for DeepPriorCBCT-reconstructed images and full-view clinical reconstructions for surgical guidance, with consistently low inter-rater agreement (Kappa <0.2) across centers. These outcomes indicate that CBCT reconstruction at 1/6 sparse sampling satisfies clinical requirements for guiding thoracic puncture procedures. The validation also moves beyond the conventional paradigm of AI algorithm development, which often prioritizes quantitative metrics over clinical relevance, by confirming practical utility through multicenter physician-driven image selection.

Despite these advances, several limitations remain: the dataset was restricted to thoracic lesions, highlighting the need for validation in other anatomical regions (e.g., head, abdomen); the prospective sample size was limited to 138 patients, necessitating larger cohorts to establish long-term clinical benefit in reducing radiation exposure; this study did not conduct Randomized Controlled Trial (RCT) to verify whether CBCB affects surgical efficiency and complications, and we will continue to carry out RCT for such verification in subsequent research.

By progressing through sequential stages of retrospective modeling, multicenter validation, and prospective clinical testing, this study developed a low-dose DeepPriorCBCT framework that reduces patient radiation while maintaining diagnostic image quality comparable to standard full-dose CBCT. The approach offers a clinically viable alternative to conventional high-dose CBCT-guided thoracic puncture procedures, with substantial potential to improve patient safety and procedural outcomes.

## MATERIALS AND METHODS

### Overview of the study

**Figure 1** presents the overall study design, which involved a multicenter early-stage clinical validation. Step 1 entailed developing a deep learning model for low-dose CBCT reconstruction using retrospectively collected pre-CBCT images from 4102 patients. Step 2 involved selecting the optimal low-dose reconstruction model based on four quantitative image quality metrics, followed by independent assessments from 11 physicians. Step 3 prospectively enrolled 138 patients undergoing CBCT-guided thoracic puncture to evaluate the clinical utility of the model through undersampling and reconstruction, with image evaluations performed by 25 physicians across 5 medical centers. Ethical approval was obtained from the institutional review board; informed consent was waived for the retrospective analysis, whereas written consent was secured from all participants in the COOS study.

### Patients

Pre-construction CBCT data were retrospectively collected from 4169 patients across 12 centers (listed in **Supplementary Appendix 1**), of which 4102 video sequences met inclusion criteria and were analyzed. Eligible patients were adults ($\geq$18 years) who had undergone CBCT-guided PTB for the relevant diseases. Exclusion was based on technical issues, such as incomplete uploads or failure in image cropping (**Supplementary Figure 2**).

Within the COOS study, 138 patients from 5 medical centers (centers 1–5) were prospectively enrolled. Inclusion required: (1) undergoing CBCT-guided percutaneous thoracic biopsy during the study period; (2) availability of complete clinical records containing demographic data and medical history; and (3) provision of written informed consent by the patient or a legal representative. Exclusion applied to: (1) patients younger than 18 years; (2) those with severe psychiatric disorders or cognitive impairment precluding study comprehension and consent; (3) pregnant women, due to radiosensitivity of the fetus and pregnancy-related imaging alterations; and (4) concurrent participation in other studies that could interfere with outcomes (**Figure 4**). All enrolled patients received two scans during CBCT-guided puncture: one full-sampling scan and one 1/6-view undersampling scan. Full-view CBCT data were also manually reduced to 1/6 views to generate reference undersampled sequences.

**Reconstruction model**

To fully exploit large-scale data and construct a reconstruction model with strong generalization across heterogeneous sparse-view conditions, a three-stage framework combining neural discrete representation learning with prior-guided reconstruction was developed. The first stage, High-Dose Prior Learning, applies an autoencoder structure—consisting of an encoder, a discrete codebook, and a decoder—to perform self-reconstruction on full-view results. This process enables the extraction of structural information from high-dose data, which is preserved in the codebook as a reliable prior. The second stage, Low-Dose Reconstruction, retains the autoencoder while introducing a classifier that links features derived from sparse-view inputs to their corresponding high-dose representations in the codebook, enabling prior-driven reconstruction of accurate outputs. The third stage, Feature Fusion, improves reconstruction fidelity by integrating multi-resolution intermediate features from both encoder and decoder, thereby refining fine-grained details while preserving global structural integrity (**Supplementary Figure 3**). The structural details of these two sub-networks are described below:

1. **Overview**

To construct a reconstruction framework capable of adapting to diverse sparse-view conditions while exploiting large-scale data, a three-stage strategy was designed that integrates neural discrete representation learning with prior-guided reconstruction. The first stage, High-Dose Prior Learning, employs an autoencoder consisting of an encoder, a discrete codebook, and a decoder to perform self-reconstruction on full-view outputs. Through this process, structural patterns from high-dose data are captured and encoded into the codebook as priors. The second stage, Low-Dose Reconstruction, preserves the autoencoder backbone while introducing a classifier that maps sparse-view (low-dose) features to their corresponding high-dose representations within the codebook, thereby enabling prior-constrained reconstruction with improved fidelity. The final stage, Feature Fusion, enhances output authenticity by integrating multi-resolution intermediate features from both encoder and decoder, allowing refinement of fine-grained details while preserving overall anatomical coherence.

2. **High-Dose Learning stage**

Building on the concept of neural discrete representation learning(*40*), an autoencoder structure comprising an encoder, a codebook, and a decoder was employed to capture prior knowledge inherent in extensive full-view X-ray projections through self-reconstruction of their corresponding full-view reconstruction outputs. Specifically, the full-view X-ray projections are first reconstructed into a CT volume $\mathbf{I}_F \in \mathbb{R}^{H \times W \times D}$ using the FDK algorithm(*13*). This volume is then transformed into a full-view feature volume $\mathbf{Z}_F \in \mathbb{R}^{h \times w \times d \times c}$ through a full-view encoder According to the vector $\mathbf{E}_F$ quantization mechanism(*40*), each voxel in the feature volume $\mathbf{Z}_F$ is replaced by the nearest code vector from a learnable codebook $\mathrm{C} = \{\mathbf{c}_n \in \mathbb{R}^c\}_{n=1}^N$, where $K$ denotes the codebook size. As a result, the quantized feature volume $\tilde{\mathbf{Z}}_F \in \mathbb{R}^{h \times w \times d \times c}$ is obtained. This process can be $\tilde{\mathbf{z}}_F^{(i,j,k)} = \arg\min_{\mathbf{c}_n \in \mathrm{C}} \left\| \mathbf{z}_F^{(i,j,k)} - \mathbf{c}_n \right\|_2$ expressed as Finally, the full-view decoder $\mathbf{D}_F$ reconstructs the CT image by decoding the quantized feature volume $\tilde{\mathbf{Z}}_F$ to CT volume $\mathbf{I}_F$.

At this stage, the encoder, decoder, and codebook are jointly trained. The overall loss, denoted as $L_{S1}$, formulated as follows:

$$L_{S1} = L_{VQ} + \lambda_{Adv} L_{Adv} + \lambda_P L_P \tag{1}$$

where $L_{VQ}$, $L_{Adv}$ and $L_P$ are vector quantization loss(*40*), adversarial loss(*41*) and perceptual loss(*42, 43*), respectively. $L_{VQ}$ is defined as:

$$L_{VQ} = \| \mathbf{I}_F - \tilde{\mathbf{I}}_F \|_2 + \| sg[\mathbf{Z}_F] - \tilde{\mathbf{Z}}_F \|_2 + \| sg[\tilde{\mathbf{Z}}_F] - \mathbf{Z}_F \|_2, \tag{2}$$

where $sg[\cdot]$ stands for stop-gradient operation. This loss function minimizes the disparity between the input CT volume $\mathbf{I}_F$ and the reconstructed CT volume $\tilde{\mathbf{I}}_F$ while aligning encoded features $E(\mathbf{x})$ and feature codes $\tilde{\mathbf{z}}$ searched from codebook $C$.

$L_{Adv}$ incorporates a discriminator $D$ to encourage more realistic reconstructions:

$$L_{Adv} = \log D(\mathbf{I}_F) + \log(1 - D(\tilde{\mathbf{I}}_F)). \tag{3}$$

$L_P$ is the perceptual loss employed to maintain high perceptual quality during vector quantization. We use 2D pre-trained VGG-16 network(*44, 45*), denoted as $\Phi$, as the feature extractor. For 3D volume, we randomly select three orthogonal 2D planes from input data $\{\mathbf{p}_m\}_{m=1}^3 \subset \mathbf{I}_F$ and the corresponding reconstructed planes $\{\tilde{\mathbf{p}}_m\}_{m=1}^3 \subset \tilde{\mathbf{I}}_F$ at the same position to compute perceptual loss:

$$L_P = \sum_{m=1}^{3} \| \Phi(\mathbf{p}_m) - \Phi(\tilde{\mathbf{p}}_m) \|_2, \tag{4}$$

### 3. Low-Dose Reconstruction Stage

The objective is to transform sparse-view X-ray projections into CBCT images free of artifacts while preserving anatomical fidelity and fine structural details. In the initial step, we first apply the FDK algorithm to obtain the sparse-view CT volume $\mathbf{I}_S \in \mathbb{R}^{H \times W \times D}$. This volume is then encoded into a sparse-view feature volume $\mathbf{Z}_S \in \mathbb{R}^{h \times w \times d \times c}$ using a sparse-view encoder $\mathbf{E}_S$ initialized from the full-view encoder $\mathbf{E}_F$. However, in this stage, nearest-neighbor search cannot be directly applied to quantize each voxel in $\mathbf{Z}_S$ using the full-view codebook $\mathbf{C}$, due to the significant distributional discrepancy between sparse-view and full-view features. Following the strategy proposed in(*46*), we introduce a code classifier $\phi_\theta$, composed of a series of convolutional layers. The classifier takes each code feature $\mathbf{Z}^{(i,j,k)}$ as input and predicts a probability sequence $\mathbf{P}^{(i,j,k)} = \{\hat{p}_n^{(i,j,k)} \in [0,1]\}_{n=1}^{N}$, where each element $\hat{p}_n^{(i,j,k)}$ denotes the probability that $\mathbf{Z}_S^{(i,j,k)}$ corresponds to the n-th quantized code $\mathbf{c}_n \in \mathbf{C}$.

As a result, the quantized feature volume $\tilde{\mathbf{Z}}_S$ is obtained by selecting the most probable code for each code feature, defined as:

$$\tilde{\mathbf{Z}}_S^{(i,j,k)} = \mathbf{c}_{\arg\max_n \hat{p}_n^{(i,j,k)}} \tag{5}$$

where $\hat{p}_n^{(i,j,k)}$ denotes the predicted probability of replacing $\mathbf{Z}_S^{(i,j,k)}$ the with the *n*-th code $\mathbf{c}_n \in \mathbf{C}$.

During this stage, the full-view decoder $\mathbf{D}_F$ and the codebook $\mathbf{C}$ are kept fixed, while the sparse-view encoder $\mathbf{E}_S$ and the code classifier $\phi_\theta$ are fine-tuned. To push the sparse-view feature volume $\mathbf{Z}_S$ closer to the full-view feature volume $\mathbf{Z}_F$, the feature loss is performed:

$$\mathrm{L}_{feature} = \| \mathbf{Z}_S - \mathbf{Z}_F \|_2 \tag{6}$$

where, $\mathbf{Z}_S = \mathbf{E}_S(\mathbf{I}_S)$ and $\mathbf{Z}_F = \mathbf{E}_F(\mathbf{I}_F)$ is the per-trained encoder in the first stage. Moreover, the classification loss (cross-entropy loss) is required measures the discrepancy between the predicted probabilities and the true labels for each code:

$$\mathrm{L}_{CE} = -\sum_{i=1}^{h}\sum_{j=1}^{w}\sum_{k=1}^{d}\sum_{n=1}^{N} \mathbf{1}(y^{(i,j,k)} = v)\log \hat{p}_v^{(i,j,k)}, \tag{7}$$

where $y^{(i,j,k)}$ is the true label of the code at position $(i,j,k)$, obtained from the pretrained autoencoder in the first stage.

The overall loss in this stage is formulated as:

$$L_{S2} = L_{feature} + L_{CE} \tag{8}$$

## 4. Feature Fusion Stage

To improve reconstruction fidelity, we fuse the multi-resolution intermediate features from the encoder and decoder, denoted as $F_E$ and $F_D$, respectively. The fused feature is given by

$$F_D = F_D + \xi_\theta(\text{concat}(F_E, F_D)), \tag{9}$$

where, $\xi_\theta$ denotes a stack of convolutions that learns to fuse the concatenated features.

In this stage, we train the convolutional fusion module $\xi_\theta$ and fine-tune the sparse-view encoder $E_S$, while keeping all other components fixed. The training objective follows the structure of the first stage but omits terms associated with codebook optimization. Specifically, it consists of the reconstruction loss, perceptual loss, and adversarial loss, formulated as:

$$L_{S3} = \|I_F - \hat{I}_F\|_2 + \lambda_{Adv}L_{Adv} + \lambda_P L_P, \tag{10}$$

where $I_F$ is the ground-truth full-view reconstruction and $\hat{I}_F$ is the final output.

## 5. Implementation Details

The framework was implemented in PyTorch(*47*). Training employed the Adam optimizer(*48*) to minimize the loss function and update network parameters. CBCT images were reconstructed from X-ray inputs using the FDK algorithm provided by the ASTRA Toolbox(*49*). All images were resampled to isotropic voxel spacing of 0.5 × 0.5 × 0.5, and intensities were normalized to [−1, 1] via min-max scaling. Random cropping was applied across all three stages during training with a patch size of 128 × 128 × 128. The learning rate was fixed at 0.0001, with batch sizes of 3 in Stages 1 and 3 and 4 in Stage 2. All training and inference procedures were executed on a single

NVIDIA Tesla A100 GPU with 80 GB memory.

**Ablation experiment**

The framework comprises three stages, and its effectiveness was assessed through ablation analysis. Three model variants were trained: (1) the complete framework (DeepPriorCBCT), (2) the framework excluding both Stage 2 and Stage 3 (ours w/o S2,3), and (3) the framework excluding only Stage 3 (ours w/o S3). An additional configuration with only Stage 2 removed was not included, as Stage 3 specifically refines feature fusion by adjusting the decoder and relies on the encoder and code classifier pre-trained in Stage 2. All models were trained under an identical data partitioning scheme (including 1015 test cases), with sparse-view X-ray projections generated by downsampling full views at a 1/6 ratio.

**Comparison experiment**

Comprehensive comparative experiments were conducted to assess the performance of the proposed method against representative baselines from both analytical and learning-based categories. The FDK algorithm, widely used in clinical CBCT reconstruction, was included as the reference standard despite its well-documented susceptibility to streak artifacts under sparse-view or limited-angle conditions(*13*). Within the deep learning domain, several established approaches were evaluated. FBPConvNet(*50*), a convolutional neural network originally designed to map artifact-corrupted reconstructions generated by filtered back-projection (FBP) to artifact-free images through paired supervision, was examined. RED-CNN(*51*), a residual encoder–decoder framework incorporating skip connections to strengthen information propagation and enhance denoising by learning residual mappings, was also included. In addition, CycleGAN(*52*), an adversarial translation network employing unpaired image-to-image learning with cycle-consistency constraints, was assessed for its ability to produce visually realistic reconstructions without paired datasets. Notably, as FBPConvNet was initially tailored for fan-beam CT, slight modifications were introduced to adapt its structure to CBCT geometry, with the adapted version denoted as FDKConvNet.

**Subjective and objective assessments**

**Subjective assessments:** Subjective evaluations were conducted by 11 physicians (6 interventionalists and 5 radiologists) with 5–15 years of experience, who independently rated the images using a single-blind protocol in which the evaluators were blinded to image type in the retrospective dataset. In addition, five radiologists with 10–15 years of experience independently assessed image quality and lesion visualization using a 5-point Likert scale. A panel of 25 interventionalists (five from each center, 8–15 years of procedural experience) further compared conventional full-view reconstructions with DeepPriorCBCT images in the prospective dataset. All evaluators remained blinded to the reconstruction method and image source. In parallel, objective metrics were calculated for three groups: CBCT reconstructed with 1/6 views using the clinical method, CBCT reconstructed with DeepPriorCBCT, and CBCT reconstructed with the clinical full-view method.

For this analysis, CBCT images from 100 patients in the retrospective dataset and 138 patients in the prospective dataset were randomly selected, and overall image quality as well as lesion diagnosis were assessed using a 5-point Likert scale. A score of 1 represented very poor quality with image interpretation not feasible, whereas a score of 5 denoted excellent quality without hindrance to diagnostic assessment. Detailed scoring criteria are provided in **Supplementary Table 4**.

**Objective assessments:** To quantitatively assess the accuracy of reconstructed CBCT images, PSNR and SSIM were calculated.

**Outcomes**

The study outcomes comprised algorithm-derived and manual assessment metrics. Algorithm-derived outcomes included PSNR and SSIM, where PSNR quantifies the ratio of maximum signal power to noise power, and SSIM evaluates the similarity between reference and reconstructed images by comparing structural information. Manual assessments encompassed physician-assigned scores for image and lesion quality as well as the accuracy in distinguishing authentic from false images.

**Statistical analysis**

Continuous variables were expressed as mean ± standard deviation or interquartile range, with Student's t-test or Mann–Whitney U-test applied according to distribution characteristics. Paired t-tests were used for within-subject comparisons. Categorical variables were analyzed using chi-square or Fisher's exact test. Agreement in Visual Turing Tests among physicians was quantified using the weighted kappa coefficient (κ) with linear weighting. Interpretation of κ followed conventional thresholds: poor (<0.20), fair (0.20–0.40), moderate (0.40–0.60), good (0.60–0.80), and excellent (>0.80). The Kendall coefficient of concordance (W) was applied to measure inter-observer consistency in image and lesion scoring across 6 interventionalists and 5 radiologist readers, with agreement classified as poor (W<0.20), fair (0.20≤W<0.40), moderate (0.40≤W<0.60), strong (0.60≤W<0.80), or super (0.80≤W<1.0).

Paired T-tests for non-inferiority were applied to determine the prospective sample size, using overall image quality manual scoring as the basis. In a retrospective assessment of 100 patients, the manual score for the clinically used full-sample reconstruction method was 3.94 ± 0.24, compared with 3.91 ± 0.28 for the DeepPriorCBCT method with 1/6 sampled reconstruction. Parameters were specified as follows: significance level 0.9, one-sided test power 0.025, non-inferiority margin 0.05 (Determine based on the differences considered by the evaluating physicians), and an expected paired mean difference of 0.02. The standard deviation of score differences between the two reconstruction approaches was 0.23. From these values, the required sample size was calculated as 116 cases, considering that 20% of the patients may be censored, and the study is expected to enroll 140 patients. The formula applied was:

$$N = \left(\frac{(Z_{1-\alpha}+Z_{1-\beta})*\sigma_d}{\delta+\mu_d}\right)^2 \tag{11}$$

Where α represents the significance level, β the test power, $\sigma_d$ the standard deviation of the paired difference, δ the non-inferiority margin, and $\mu_d$ the expected mean paired difference.

In this study, P-values < 0.05 were regarded as statistically significant. Statistical analyses were performed using R version 4.3.1, PASS 2021 ( Matlab version R2022b (MathWorks, Natick, MA, USA), and Python version 3.10 (Python Software Foundation, Beaverton, OR, USA).

**Supplementary Materials**

Materials and Methods

Tables S1 to S4 for multiple supplementary tables

Fig S1 to S3 for multiple supplementary figures

**References**

1. F. Bray, M. Laversanne, H. Sung, J. Ferlay, R. L. Siegel, I. Soerjomataram, A. Jemal, Global cancer statistics 2022: GLOBOCAN estimates of incidence and mortality worldwide for 36 cancers in 185 countries. *CA: a cancer journal for clinicians* **74**, 229-263 (2024).
2. S. J. Adams, E. Stone, D. R. Baldwin, R. Vliegenthart, P. Lee, F. J. Fintelmann, Lung cancer screening. *Lancet (London, England)* **401**, 390-408 (2023).
3. P. J. Mazzone, L. Lam, Evaluating the Patient With a Pulmonary Nodule: A Review. *Jama* **327**, 264-273 (2022).
4. J. A. Howington, M. G. Blum, A. C. Chang, A. A. Balekian, S. C. Murthy, Treatment of stage I and II non-small cell lung cancer: Diagnosis and management of lung cancer, 3rd ed: American College of Chest Physicians evidence-based clinical practice guidelines. *Chest* **143**, e278S-e313S (2013).
5. R. Lencioni, L. Crocetti, R. Cioni, R. Suh, D. Glenn, D. Regge, T. Helmberger, A. R. Gillams, A. Frilling, M. Ambrogi, C. Bartolozzi, A. Mussi, Response to radiofrequency ablation of pulmonary tumours: a prospective, intention-to-treat, multicentre clinical trial (the RAPTURE study). *The Lancet. Oncology* **9**, 621-628 (2008).
6. S. M. Schwartz, Epidemiology of Cancer. *Clin Chem* **70**, 140-149 (2024).
7. H. Li, H. Li, M. Zhang, C. Huang, X. Zhou, Direct imaging of pulmonary gas exchange with hyperpolarized xenon MRI. *The Innovation* **5**,  (2024).
8. D. B. Richardson, K. Leuraud, D. Laurier, M. Gillies, R. Haylock, K. Kelly-Reif, S. Bertke, R. D. Daniels, I. Thierry-Chef, M. Moissonnier, A. Kesminiene, M. K. Schubauer-Berigan, Cancer mortality after low dose exposure to ionising radiation in workers in France, the United Kingdom, and the United States (INWORKS): cohort study. *BMJ (Clinical research ed.)* **382**, e074520 (2023).
9. M. Bosch de Basea, I. Thierry-Chef, R. Harbron, M. Hauptmann, G. Byrnes, M. O. Bernier, L. Le Cornet, J. Dabin, G. Ferro, T. S. Istad, A. Jahnen, C. Lee, C. Maccia, F. Malchair, H. Olerud, S. L. Simon, J. Figuerola, A. Peiro, H. Engels, C. Johansen, M. Blettner, M. Kaijser, K. Kjaerheim, A. Berrington de Gonzalez, N. Journy, J. M. Meulepas, M. Moissonnier, A. Nordenskjold, R. Pokora, C. Ronckers, J. Schüz, A. Kesminiene, E. Cardis, Risk of hematological malignancies from CT radiation exposure in children, adolescents and young adults. *Nat Med* **29**, 3111-3119 (2023).
10. M. Hauptmann, G. Byrnes, E. Cardis, M. O. Bernier, M. Blettner, J. Dabin, H. Engels, T. S. Istad, C. Johansen, M. Kaijser, K. Kjaerheim, N. Journy, J. M. Meulepas, M. Moissonnier, C. Ronckers, I. Thierry-Chef, L. Le Cornet, A. Jahnen, R. Pokora, M. Bosch de Basea, J. Figuerola, C. Maccia, A. Nordenskjold, R. W. Harbron, C. Lee, S. L. Simon, A. Berrington de Gonzalez, J. Schüz, A. Kesminiene, Brain cancer after radiation exposure from CT examinations of children and young adults: results from the EPI-CT cohort study. *The Lancet. Oncology* **24**, 45-53 (2023).

11. M. P. Little, T. V. Azizova, D. B. Richardson, S. Tapio, M. O. Bernier, M. Kreuzer, F. A. Cucinotta, D. Bazyka, V. Chumak, V. K. Ivanov, L. H. S. Veiga, A. Livinski, K. Abalo, L. B. Zablotska, A. J. Einstein, N. Hamada, Ionising radiation and cardiovascular disease: systematic review and meta-analysis. *BMJ (Clinical research ed.)* **380**, e072924 (2023).
12. R. Smith-Bindman, P. W. Chu, H. Azman Firdaus, C. Stewart, M. Malekhedayat, S. Alber, W. E. Bolch, M. Mahendra, A. Berrington de González, D. L. Miglioretti, Projected Lifetime Cancer Risks From Current Computed Tomography Imaging. *JAMA Intern Med*, (2025).
13. L. A. Feldkamp, L. C. Davis, J. W. Kress, Practical cone-beam algorithm. *Journal of the Optical Society of America A* **1**, 612-619 (1984).
14. W. Chen, Y. Li, N. Yuan, J. Qi, B. A. Dyer, L. Sensoy, S. H. Benedict, L. Shang, S. Rao, Y. Rong, Clinical enhancement in AI-based post-processed fast-scan low-dose CBCT for head and neck adaptive radiotherapy. *Frontiers in artificial intelligence* **3**, 614384 (2021).
15. A. J. Olch, P. Alaei, How low can you go? A CBCT dose reduction study. *Journal of Applied Clinical Medical Physics* **22**, 85-89 (2021).
16. Y. Chan, M. Li, K. Parodi, C. Belka, G. Landry, C. Kurz, Feasibility of CycleGAN enhanced low dose CBCT imaging for prostate radiotherapy dose calculation. *Physics in Medicine & Biology* **68**, 105014 (2023).
17. S. Kim, H. Lee, W.-J. Yi, M. K. Cho, in *Medical Imaging 2024: Physics of Medical Imaging*. (SPIE, 2024), vol. 12925, pp. 681-687.
18. Y. Lin, Z. Luo, W. Zhao, X. Li, in *International Conference on Medical Image Computing and Computer-Assisted Intervention*. (Springer, 2023), pp. 13-23.
19. Y. Lin, H. Wang, J. Chen, X. Li, in *International Conference on Medical Image Computing and Computer-Assisted Intervention*. (Springer, 2024), pp. 425-435.
20. S. Zhang, Y. Liu, D. Li, Artificial intelligence-and pediatric CBCT-aided developmental recognition and personalized treatment of periapical diseases. *The Innova*, (2025).
21. H. Zhao, Z. Xu, L. Chen, L. Wu, Z. Cui, J. Ma, T. Sun, Y. Lei, N. Wang, H. Hu, Y. Tan, W. Lu, W. Yang, K. Liao, G. Teng, X. Liang, Y. Li, C. Feng, T. Nie, X. Han, D. Xiang, C. Majoie, W. H. van Zwam, A. van der Lugt, P. M. van der Sluijs, T. van Walsum, Y. Feng, G. Liu, Y. Huang, W. Liu, X. Kan, R. Su, W. Zhang, X. Wang, C. Zheng, Large-scale pretrained frame generative model enables real-time low-dose DSA imaging: An AI system development and multi-center validation study. *Med* **6**, 100497 (2025).
22. G. Zhang, Y. Zhu, H. Wang, Y. Chen, G. Wu, L. Wang, in *Proceedings of the IEEE/CVF Conference on Computer Vision and Pattern Recognition*. (2023), pp. 5682-5692.
23. F. Reda, J. Kontkanen, E. Tabellion, D. Sun, C. Pantofaru, B. Curless, in *European Conference on Computer Vision*. (Springer, 2022), pp. 250-266.
24. H. Zhao, Z. Zhou, F. Wu, D. Xiang, H. Zhao, W. Zhang, L. Li, Z. Li, J. Huang, H. Hu, C. Liu, T. Wang, W. Liu, J. Ma, F. Yang, X. Wang, C. Zheng, Self-supervised learning enables 3D digital subtraction angiography reconstruction from ultra-sparse 2D projection views: A multicenter study. *Cell reports. Medicine* **3**, 100775 (2022).
25. Z. Xu, H. Zhao, W. Liu, X. Wang, in *Proceedings of the AAAI Conference on Artificial Intelligence*. (2025), vol. 39, pp. 28530-28538.
26. R. Smyth, E. Billatos, Novel Strategies for Lung Cancer Interventional Diagnostics. *J Clin Med* **13**, (2024).
27. H. Kim, J.-S. Choi, Y. Lee, Assessment of Feldkamp-Davis-Kress Reconstruction Parameters in

Overall Image Quality in Cone Beam Computed Tomography. *Applied Sciences* **14**, 7058 (2024).

28. K. Tsiklakis, C. Donta, S. Gavala, K. Karayianni, V. Kamenopoulou, C. J. Hourdakis, Dose reduction in maxillofacial imaging using low dose Cone Beam CT. *European journal of radiology* **56**, 413-417 (2005).
29. C. Wang, K. Shang, H. Zhang, Q. Li, S. K. Zhou, in *International workshop on machine learning for medical image reconstruction*. (Springer, 2022), pp. 84-94.
30. R. Li, Q. Li, H. Wang, S. Li, J. Zhao, Q. Yan, L. Wang, DDPTransformer: Dual-domain with parallel transformer network for sparse view CT image reconstruction. *IEEE Transactions on Computational Imaging* **8**, 1101-1116 (2022).
31. R. Zha, Y. Zhang, H. Li, in *International Conference on Medical Image Computing and Computer-Assisted Intervention*. (Springer, 2022), pp. 442-452.
32. Y. Cai, J. Wang, A. Yuille, Z. Zhou, A. Wang, in *Proceedings of the IEEE/CVF conference on computer vision and pattern recognition*. (2024), pp. 11174-11183.
33. I. Ayad, N. Larue, M. K. Nguyen, in *Proceedings of the IEEE/CVF conference on computer vision and pattern recognition*. (2024), pp. 25317-25326.
34. A. V. D. Oord, O. Vinyals, K. Kavukcuoglu, Neural Discrete Representation Learning. (2017).
35. M. Tancik, E. Weber, E. Ng, R. Li, B. Yi, T. Wang, A. Kristoffersen, J. Austin, K. Salahi, A. Ahuja, in *ACM SIGGRAPH 2023 conference proceedings*. (2023), pp. 1-12.
36. Y. Lin, J. Yang, H. Wang, X. Ding, W. Zhao, X. Li, in *Proceedings of the IEEE/CVF conference on computer vision and pattern recognition*. (2024), pp. 11205-11214.
37. Z. Liu, H. Zhao, W. Qin, Z. Zhou, X. Wang, W. Wang, X. Lai, C. Zheng, D. Shen, Z. Cui, 3D Vessel Reconstruction from Sparse-View Dynamic DSA Images via Vessel Probability Guided Attenuation Learning. *CoRR*, (2024).
38. Z. Liu, Y. Fang, C. Li, H. Wu, Y. Liu, D. Shen, Z. Cui, Geometry-aware attenuation learning for sparse-view CBCT reconstruction. *IEEE Transactions on Medical Imaging*, (2024).
39. Z. Zhang, X. Liang, X. Dong, Y. Xie, G. Cao, A sparse-view CT reconstruction method based on combination of DenseNet and deconvolution. *IEEE transactions on medical imaging* **37**, 1407-1417 (2018).
40. A. Van Den Oord, O. Vinyals, Neural discrete representation learning. *Advances in neural information processing systems* **30**, (2017).
41. P. Esser, R. Rombach, B. Ommer, in *Proceedings of the IEEE/CVF conference on computer vision and pattern recognition*. (2021), pp. 12873-12883.
42. J. Johnson, A. Alahi, L. Fei-Fei, *Perceptual Losses for Real-Time Style Transfer and Super-Resolution*. (Springer, Cham, 2016).
43. X. Yu, B. Fernando, B. Ghanem, F. Porikli, R. Hartley, Face Super-resolution Guided by Facial Component Heatmaps. *Springer, Cham*, (2018).
44. K. Simonyan, A. Zisserman, in *International Conference on Learning Representations*. (2015).
45. J. Deng, W. Dong, R. Socher, L. J. Li, K. Li, L. Fei-Fei, ImageNet: A large-scale hierarchical image database. *Proc of IEEE Computer Vision & Pattern Recognition*, 248-255 (2009).
46. S. Zhou, K. Chan, C. Li, C. C. Loy, Towards robust blind face restoration with codebook lookup transformer. *Advances in Neural Information Processing Systems* **35**, 30599-30611 (2022).
47. A. Paszke, A. Lerer, T. Killeen, L. Antiga, E. Yang, A. Tejani, L. Fang, S. Gross, J. Bradbury, Z. Lin, in *Advances in Neural Information Processing Systems 32, Volume 11 of 20: 32nd Conference on Neural Information Processing Systems (NeurIPS 2019).Vancouver(CA).8-14 December 2019*.

(2020).
48. D. P. Kingma, J. Ba, Adam: A Method for Stochastic Optimization. *Ithaca, NYarXiv.org*, (2014).
49. W. V. Aarle, W. J. Palenstijn, J. Cant, E. Janssens, J. Sijbers, Fast and flexible X-ray tomography using the ASTRA toolbox. *Optics Express* **24**, 25129-25147 (2016).
50. K. H. Jin, M. T. McCann, E. Froustey, M. Unser, Deep convolutional neural network for inverse problems in imaging. *IEEE transactions on image processing* **26**, 4509-4522 (2017).
51. H. Chen, Y. Zhang, M. K. Kalra, F. Lin, Y. Chen, P. Liao, J. Zhou, G. Wang, Low-dose CT with a residual encoder-decoder convolutional neural network. *IEEE transactions on medical imaging* **36**, 2524-2535 (2017).
52. Z. Li, J. Huang, L. Yu, Y. Chi, M. Jin, in *2019 IEEE nuclear science symposium and medical imaging conference (NSS/MIC)*. (IEEE, 2019), pp. 1-3.

**Acknowledgements:** We extend our sincere gratitude to patients who are willing to participate in the study.

**Funding:**

National Key Research and Development Program of China (2023YFC2705700)

National Natural Science Foundation of China (Grant No. 62225113, 82472070, 82402411)

Natural Science Foundation of Hubei Provincial (Grant No. 2025DJA055, 2023AFB1083)

The Key Research and Development Program of Hubei Province (Grant No. 2024BCB036).

**Author Contributions:**

Conceptualization: ZHX, DB, TGJ

Methodology: WHS, CL, ZLF, ZW, ZXY, LY

Investigation: CL, ZWH, WHS, WLX, WZM, ZW

Visualization: ZHX, DB, TGJ

Funding acquisition: ZHX, DB, TWJ, CL, XY

Project administration: ZHX, DB

Supervision: ZHX, DB, TGJ

Writing – original draft: CL, WHS, WLX, ZWH, ZCC

Writing – review & editing: ZHX, DB, TGJ

**Competing interests:** We declare that we have no known competing financial interests or personal relationships that could have appeared to influence the work reported in this paper.

**Data and materials availability:** The CBCT images from ten patients can be available from the URL (https://github.com/Mors20/DeepPriorCBCT), which can be used for the image generation. The remaining data is available for research purposes upon reasonable request from the corresponding author, Huangxuan Zhao.

**Code availability:** The codes used for model construction can be available from the URL (https://github.com/Mors20/DeepPriorCBCT).

**Figures**

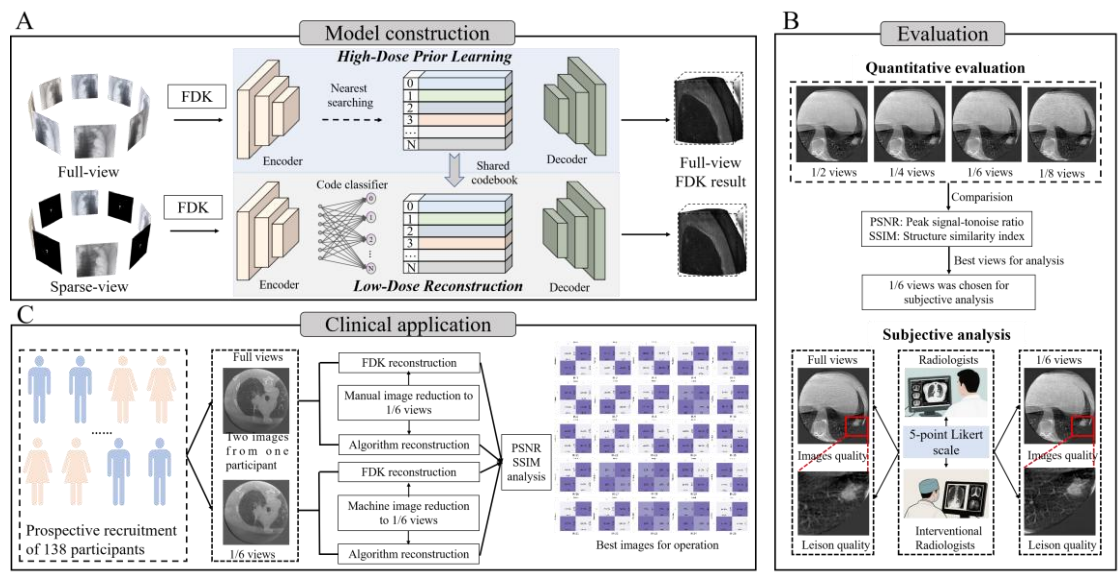

**Figure 1.** Flowchart of the study. (A) DeepPriorCBCT model construction; (B) Optimal selection of sparse image reconstruction models and subjective analysis of reconstructed images; (C) Prospective clinical application of the DeepPriorCBCT.

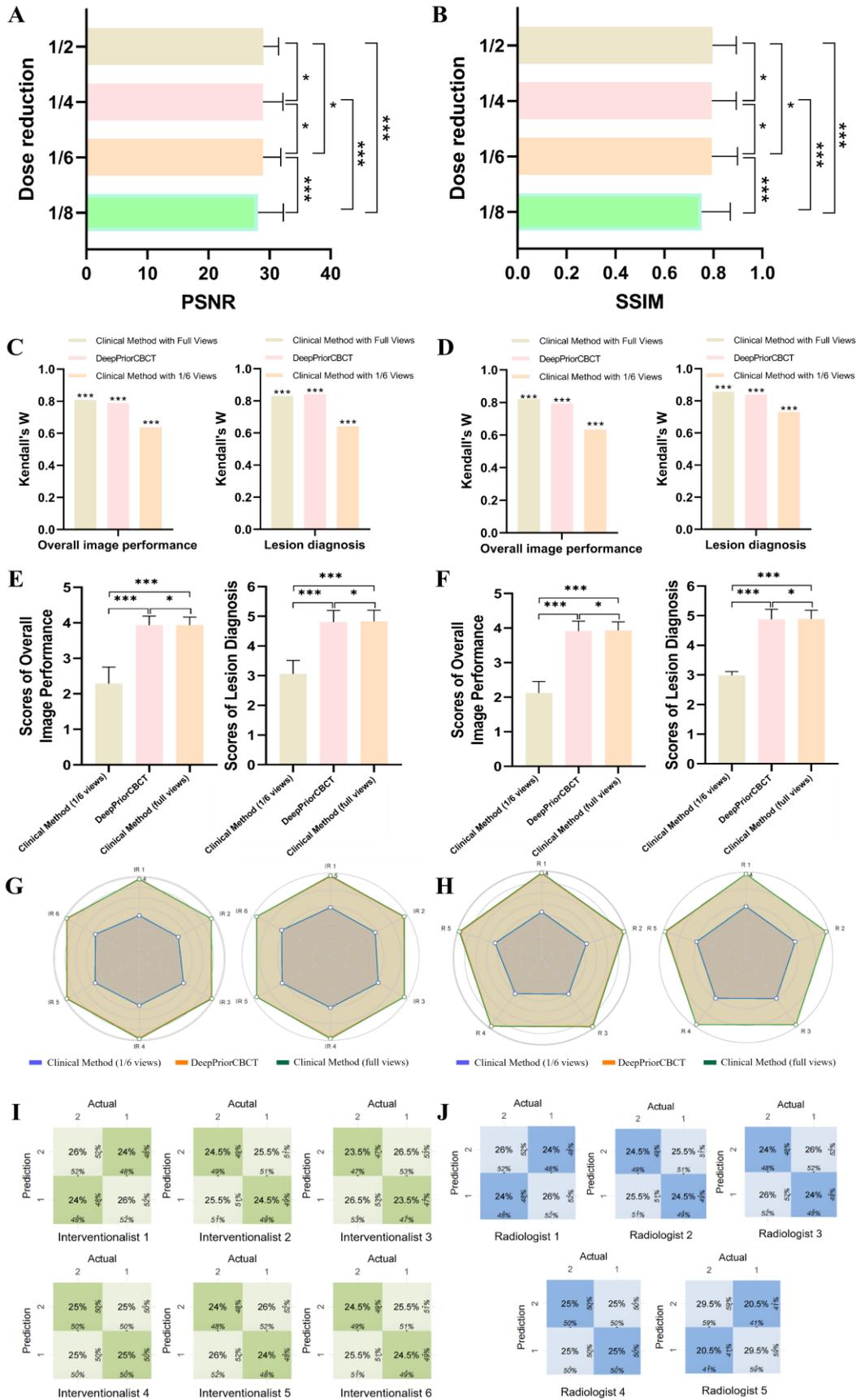

**Figure 2.** Subjective evaluation on retrospective data. (A–B) Ratios of algorithmic metrics across different views; (C) Inter-observer agreement in image and lesion

scoring among interventionalists; (D) Agreement among radiologists; (E) Comparison of image and lesion scores from interventionalists; (F) Comparison of scores from radiologists; (G) Radar plots of score distributions for interventionalists; (H) Radar plots for radiologists; (I) Confusion matrix for interventionalists in authenticity recognition; (J) Confusion matrix for radiologists. * P>0.05, *** P<0.001.

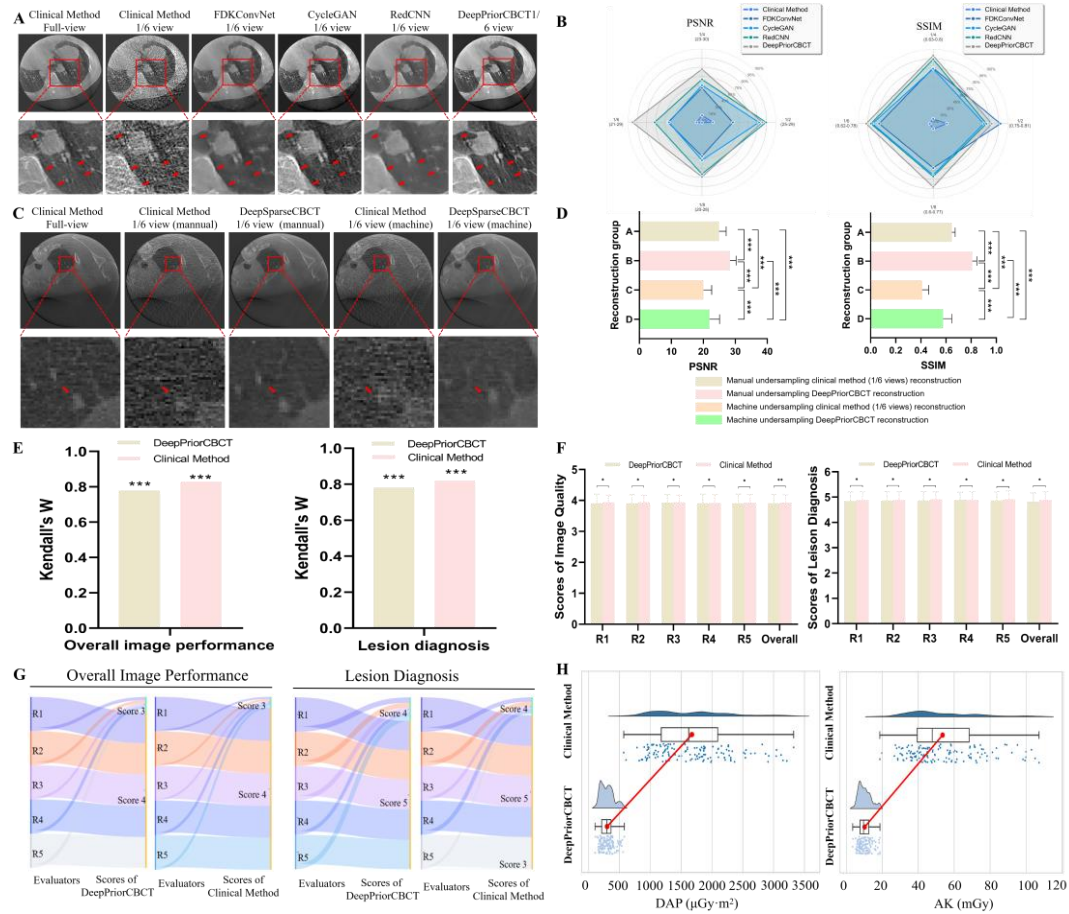

**Figure 3.** Comparative experimental and prospective study outcomes. (A) CBCT reconstructions from five models, with red arrows highlighting detail loss in four models compared with the clinically applied full-view method and DeepPriorCBCT; (B) Radar chart comparing PSNR and SSIM across reconstructions from different view numbers; (C–D) CBCT images reconstructed by various methods, with red arrows marking detail loss in three approaches relative to full-view reconstruction and DeepPriorCBCT; (E) Inter-rater consistency in image and lesion scoring among five radiologists; (F) Comparison of image and lesion scores across the five radiologists; (G) Alluvial plot illustrating the relationship between overall image quality and lesion diagnosis from five radiologists; (H) Radiation dose distribution of CBCT scans

reconstructed by DeepPriorCBCT and the clinical full-view method. * P>0.05, ** P<0.05, *** P<0.001.

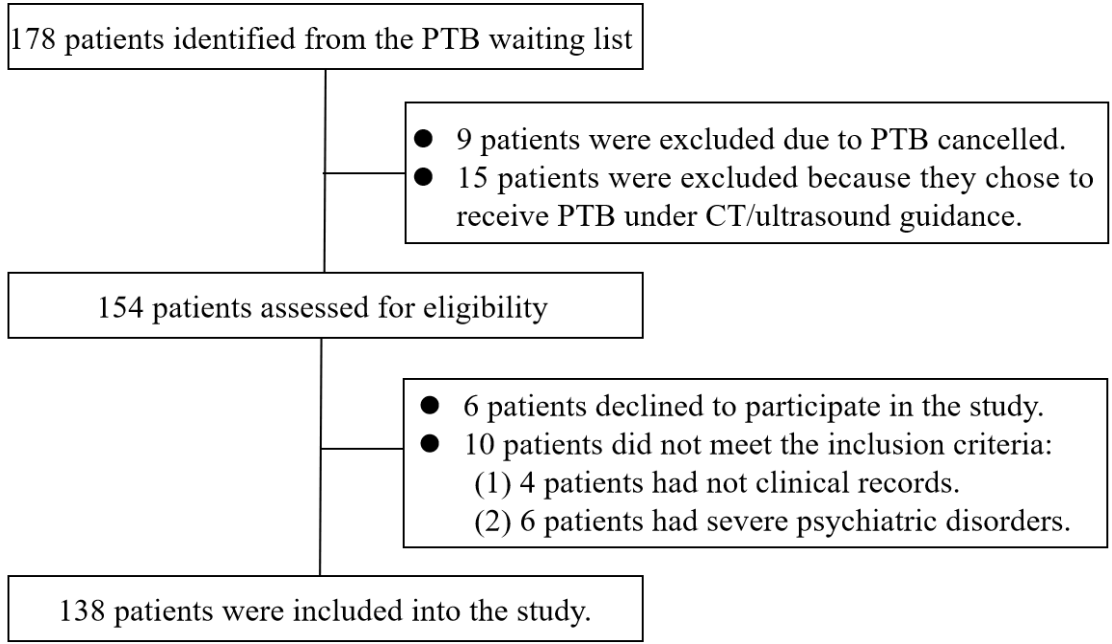

**Figure 4.** CONSORT flow diagram.

**Table 1.** Baseline characteristics of patients and data source centers

|  | Retrospective cohort | | | Prospective cohort |
| --- | --- | --- | --- | --- |
|  | Training group (n=2987) | Validation group (n=100) | Test group (n=1015) | COOS (n=138) |
| **Age（years）** | 60.8±11.5 | 62.1±12.5 | 60.6±13.2 | 63.0±10.5 |
| **Sex** | | | | |
| Male | 1983 (66.4) | 68 (68.0) | 680 (67.0) | 84 (60.9) |
| Female | 1004 (33.6) | 32 (32.0) | 335 (33.0) | 54 (39.1) |
| **BMI（kg/m2）** | 22.3±3.5 | 22.6±4.3 | 22.4±3.7 | 21.9±4.6 |
| **Lesion size（cm）** | 5.6±1.1 | 4.8±1.8 | 6.1±1.3 | 5.3±2.1 |
| **Lesion location** | | | | |
| Left lung | 1378 (46.1) | 45 (45.0) | 466 (45.9) | 66 (47.8) |
| Right Lung | 1403 (47.0) | 48 (48.0) | 478 (47.1) | 63 (45.7) |
| Mediastinum | 206 (6.9) | 7 (7.0) | 71 (7.0) | 9 (6.5) |
| | **Data source centers** | | | |
| | Center 1 (n=1048) | Center 1 (n=100) | | Center 1 (n=48) |
| | Center 2 (n=315) | | | Center 2 (n=28) |
| | Center 3 (n=227) | | | Center 3 (n=21) |
| | Center 4 (n=279) | | | Center 4 (n=23) |
| | Center 5 (n=389) | | | Center 5 (n=18) |
| | Center 6 (n=195) | | | |
| | Center 7 (n=306) | | | |
| | Center 8 (n=228) | | | |
| | | | Center 9 (n=329) | |
| | | | Center 10 (n=124) | |
| | | | Center 11 (n=296) | |
| | | | Center 12 (n=266) | |

**Neural Discrete Representation Learning for Sparse-View CBCT Reconstruction: From Algorithm Design to Prospective Multicenter Clinical Evaluation**

# Contents

**Supplementary Appendix 1.** Medical centers participating in the retrospective study

Center 1: Union Hospital, Tongji Medical College, Huazhong University of Science and Technology

Center 2: Wuhan Union West Hospital

Center 3: Wuhan Union Jinyin Lake Hospital

Center 4: The First Affiliated Hospital of Zhengzhou University

Center 5: The First Affiliated Hospital East Campus of Zhengzhou University

Center 6: The First Affiliated Hospital of Guangzhou Medical University

Center 7: Zhongda Hospital Jiangbei Campus, Medical School, Southeast University

Center 8: Union Hospital, Fujian Medical University

Center 9: Zhongda Hospital, Medical School, Southeast University

Center 10: Fujian Provincial Hospital Affiliated to Fuzhou University

Center 11: The First Affiliated Hospital of University of Science and Technology of China

Center 12: Zhongshan Hospital of Fudan University

**Supplementary Table 1.** Inter-rater agreement analysis for CBCT images visual turing tests

|  |  |  |
|---|---|---|
| **Radiologists** | Kappa | -0.087 |
|  | z-value | -3.37 |
|  | p-value | <0.001 |
| **Interventional Radiologists** | Kappa | 0.091 |
|  | z-value | 2.88 |
|  | p-value | 0.003 |

**Supplementary Table 2.** Quantitative results of the ablation study

|  | PSNR ↑ | SSIM ↑ |
|---|---|---|
| DeepPriorCBCT (w/o S2,3) | 25.8981 ± 5.5215 | 0.7227 ± 0.0075 |
| DeepPriorCBCT (w/o 3) | 27.3026 ± 10.3940 | 0.7705 ± 0.0087 |
| DeepPriorCBCT | **28.8073 ± 12.5254** | **0.7734 ± 0.0123** |

**Supplementary Table 3. Consistance analysis on the selection of optimal images for surgical guidance by 25 interventionalists**

| Readers | IR-1-25 | IR1-5 | IR6-10 | IR11-15 | IR16-20 | IR21-25 |
|---|---|---|---|---|---|---|
| Kappa | -0.024 | -0.139 | -0.148 | -0.119 | -0.131 | -0.129 |
| Z-value | -4.84 | -5.18 | -5.50 | -4.43 | -4.85 | -4.79 |
| P-value | < 0.001 | < 0.001 | < 0.001 | 0.001 | < 0.001 | < 0.001 |

**Supplementary Table 4.** Five-point Likert scale for overall image quality and lesion quality assessment

| 5-point Likert scale | Overall image quality | Lesion quality |
| --- | --- | --- |
| 1 | The image appears faulty | Unable to identify lesions |
| 2 | Severe artifacts and noise in the image | Difficult to identify the location of the lesion and unable to measure the size |
| 3 | Severe artifacts or noise in the image | Harder to identify lesion location; difficult to measure size |
| 4 | Mild artifacts or noise in the image | Location of lesions can be localized, harder to measure size |
| 5 | Image free of artifacts and noise | Location of lesions can be localized, and the size can be measured |

**Supplementary Figure 1.** 25 interventionalists choose the best guided surgical Image obfuscation matrix

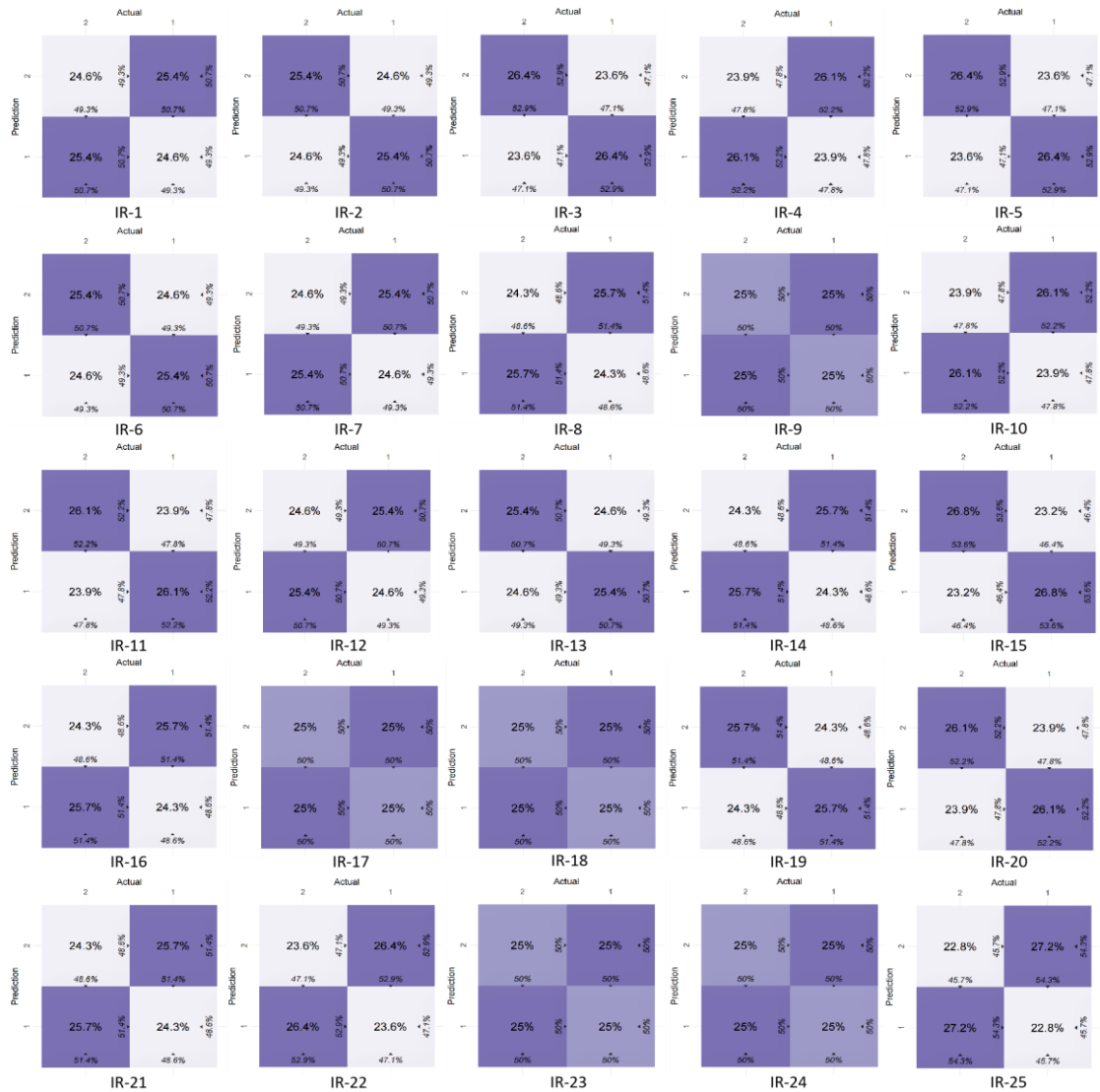

**Supplementary Figure 2.** Flowchart of patient selection in the retrospective cohort

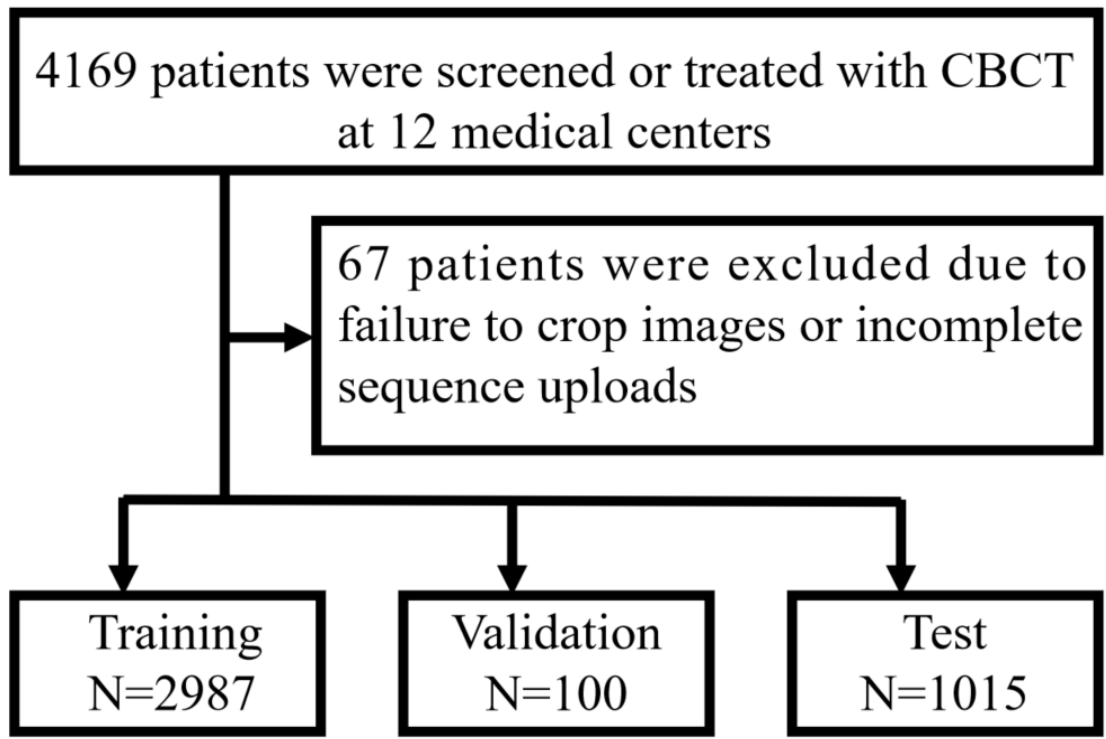

**Supplementary Figure 3.** The whole framework of the algorithm

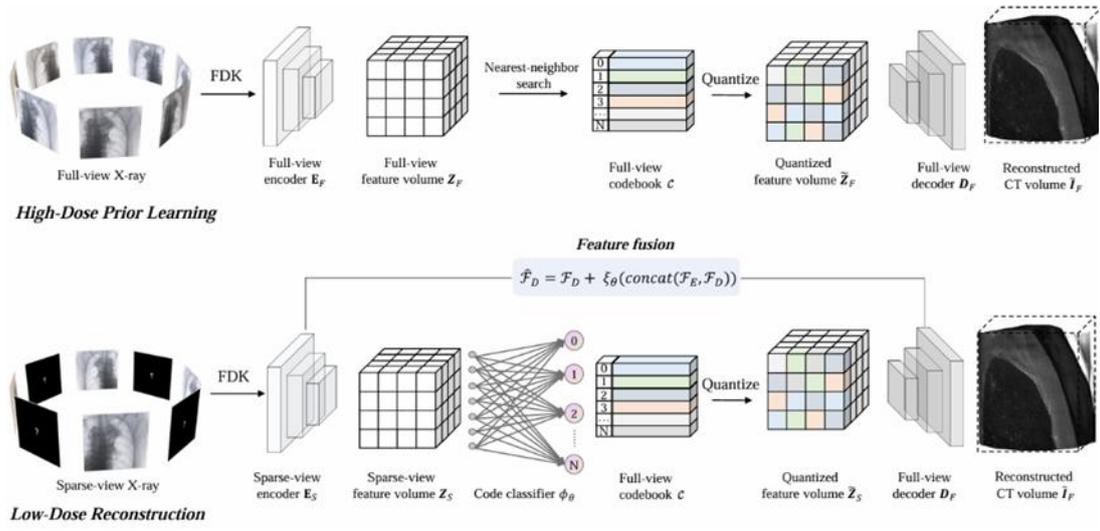